\pdfoutput=1
\documentclass[11pt, letterpaper, logo, onecolumn, copyright]{template_from_google}

\usepackage[T1]{fontenc}
\usepackage{CJKutf8} 

\usepackage{amsmath}
\usepackage{amsfonts}
\usepackage{amssymb}
\usepackage{amsthm}
\usepackage{bm} 
\usepackage{nicefrac} 

\usepackage{booktabs} 
\usepackage{enumitem} 
\usepackage{fancyhdr}
\usepackage{multirow} 
\usepackage{tabularx}
\usepackage{multicol}
\usepackage{array}
\usepackage{longtable}
\usepackage{lipsum} 
\usepackage{wrapfig} 
\usepackage{tocloft} 
\usepackage{fancybox}
\usepackage{authblk} 

\usepackage{algorithm}
\usepackage[noend]{algpseudocode}
\usepackage{algpseudocode}

\usepackage{graphicx}
\usepackage{caption}
\usepackage{subcaption}
\usepackage{capt-of} 
\usepackage[inkscapeformat=png]{svg}

\usepackage{listings, listings-rust}
\usepackage{listingsutf8}

\usepackage[dvipsnames]{xcolor}
\usepackage[most, breakable, skins]{tcolorbox}
\tcbuselibrary{listings,skins,breakable}

\usepackage{fontawesome5}
\usepackage{xspace} 

\usepackage[authoryear, sort&compress, round]{natbib}

\usepackage[dvipsnames]{xcolor} 

\usepackage{hyperref}

\definecolor{darkblue}{rgb}{0.0, 0.0, 0.6}
\definecolor{darkred}{rgb}{0.7, 0.0, 0.0}
\hypersetup{
  pdffitwindow=true,
  pdfstartview={FitH},
  pdfnewwindow=true,
  colorlinks,
  linktocpage=true,
  linkcolor=darkred,
  urlcolor=darkblue,
  citecolor=darkblue
}

\usepackage{amsmath,amsfonts,bm}

\def\eqref#1{equation~\ref{#1}}

\def\1{\bm{1}}

\def\rvc{{\mathbf{c}}}

\def\rvx{{\mathbf{x}}}

\DeclareMathAlphabet{\mathsfit}{\encodingdefault}{\sfdefault}{m}{sl}
\SetMathAlphabet{\mathsfit}{bold}{\encodingdefault}{\sfdefault}{bx}{n}

\newcommand{\E}{\mathbb{E}}

\usepackage[export]{adjustbox}
\usepackage[capitalise]{cleveref}
\crefname{equation}{Eq.}{Eqs.}
\newcommand{\bbE}{\ensuremath{\mathbb{E}}}
\newcommand{\kl}{\mathrm{KL}}

\usepackage{multirow}
\usepackage{caption}
\newcommand{\spara}[1]{\noindent\textbf{#1.}}
\usepackage{paralist}

\makeatletter

\renewcommand\paragraph{\@startsection{paragraph}{4}{\z@}%
            {-2.5ex\@plus -1ex \@minus -.25ex}%
            {1.25ex \@plus .25ex}%
            {\itshape\normalsize\bfseries}}
\makeatother

\let\cite\citep

\title{TDM-R1: Reinforcing Few-Step Diffusion Models with Non-Differentiable Reward}
\newcommand{\shorttitle}{TDM-R1}
\renewcommand{\headrulewidth}{0.4pt}

\reportnumber{} 

\renewcommand{\today}{}

\author[1]{Yihong Luo}
\author[2]{Tianyang Hu}
\author[3]{Weijian Luo}
\author[4,1]{Jing Tang}

\affil[1]{Hong Kong University of Science and Technology}
\affil[2]{The Chinese University of Hong Kong, Shenzhen}
\affil[3]{hi-Lab, Xiaohongshu Inc}
\affil[4]{Hong Kong University of Science and Technology (Guangzhou)}

\begin{abstract}
While few-step generative models have enabled powerful image and video generation at significantly lower cost, generic reinforcement learning (RL) paradigms for few-step models remain an unsolved problem. Existing RL approaches for few-step diffusion models strongly rely on back-propagating through differentiable reward models, thereby excluding the majority of important real-world reward signals, e.g., non-differentiable rewards such as humans' binary likeness, object counts, etc. 
To properly incorporate non-differentiable rewards to improve few-step generative models, we introduce TDM-R1, a novel reinforcement learning paradigm built upon a leading few-step model, Trajectory Distribution Matching (TDM). 
TDM-R1 decouples the learning process into surrogate reward learning and generator learning.
Furthermore, we developed practical methods to obtain per-step reward signals along the deterministic generation trajectory of TDM, resulting in a unified RL post-training method that significantly improves few-step models' ability with generic rewards.
We conduct extensive experiments ranging from text-rendering, visual quality, and preference alignment. All results demonstrate that TDM-R1 is a powerful reinforcement learning paradigm for few-step text-to-image models, achieving state-of-the-art reinforcement learning performances on both in-domain and out-of-domain metrics. Furthermore, TDM-R1 also scales effectively to the recent strong Z-Image model, consistently outperforming both its 100-NFE and few-step variants with only 4 NFEs. 
Project page: \url{https://github.com/Luo-Yihong/TDM-R1}.
\end{abstract}

\begin{document}

\maketitle

\vspace{-5mm}
\section{Introduction}

Achieving rapid, high-fidelity image and video generation continues to be a fundamental objective in the field of Artificial Intelligence Generated Content (AIGC). 
Recent advancements in few-step generative models, e.g., diffusion distillation \citep{luo2023diff,luo2024one,yin2023one,zhou2024score,xie2024distillation,luo2025learning}, have shown successes in ultra-fast photo-realistic images and videos generation with accelerations up to 50 times that of diffusion models. With ultra-fast generation efficiency and leading performances, few-step models are becoming a standard with large-scale serving in industry products \citep{cai2025z,flux2024}. {\textit{Despite significant gains in speed and visual fidelity, few-step models still struggle with certain challenges, e.g., precise \textit{instruction following}, complicated text rendering, and proper object positioning.}}

In recent years, reinforcement learning has shown significant ability in improving deep learning models' specialized abilities, ranging from large language models \citep{ziegler2019fine, ouyang2022training, bai2022training, shao2024deepseekmath, deepseek_r1}, standard diffusion models \citep{flowgrpo,dancegrpo}, to few-step generative models \citep{luo2024diff,luo2025reward,luo2025adding,ren2024hyper}. It is expected that proper RL paradigms are capable of addressing the mentioned challenges of few-step image or video generation. 

\begin{figure*}[!t]
    \centering
    \includegraphics[width=0.99\linewidth]{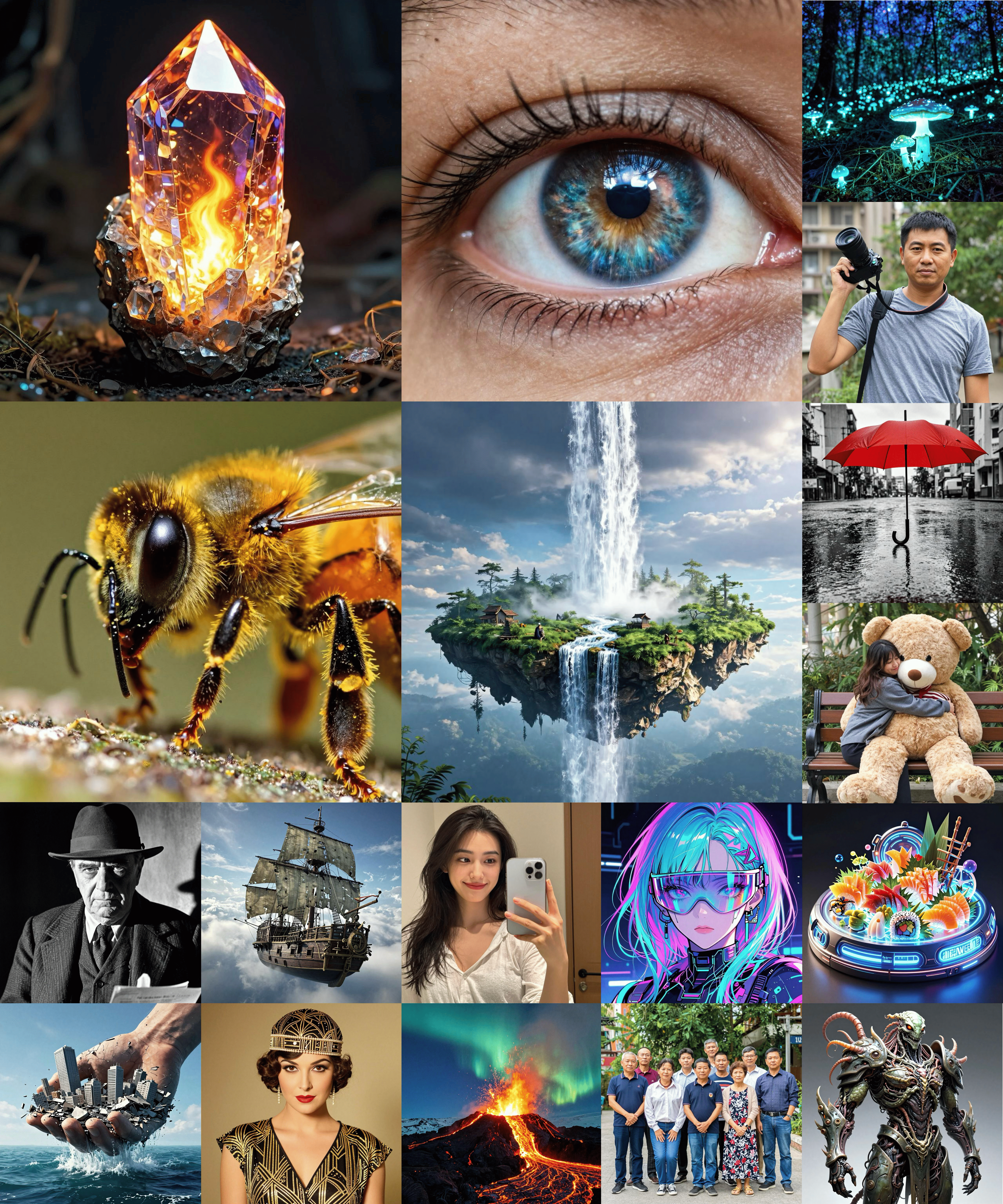}
    \caption{
    Samples generated by \textbf{TDM-R1} using only 4 NFEs, obtained by reinforcing the recent powerful Z-Image model~\cite{zimage}.
    }
    \label{fig:visual_compare}
    \vspace{-4mm}
\end{figure*}

\begin{figure}
    \centering
    \includegraphics[width=0.95\linewidth]{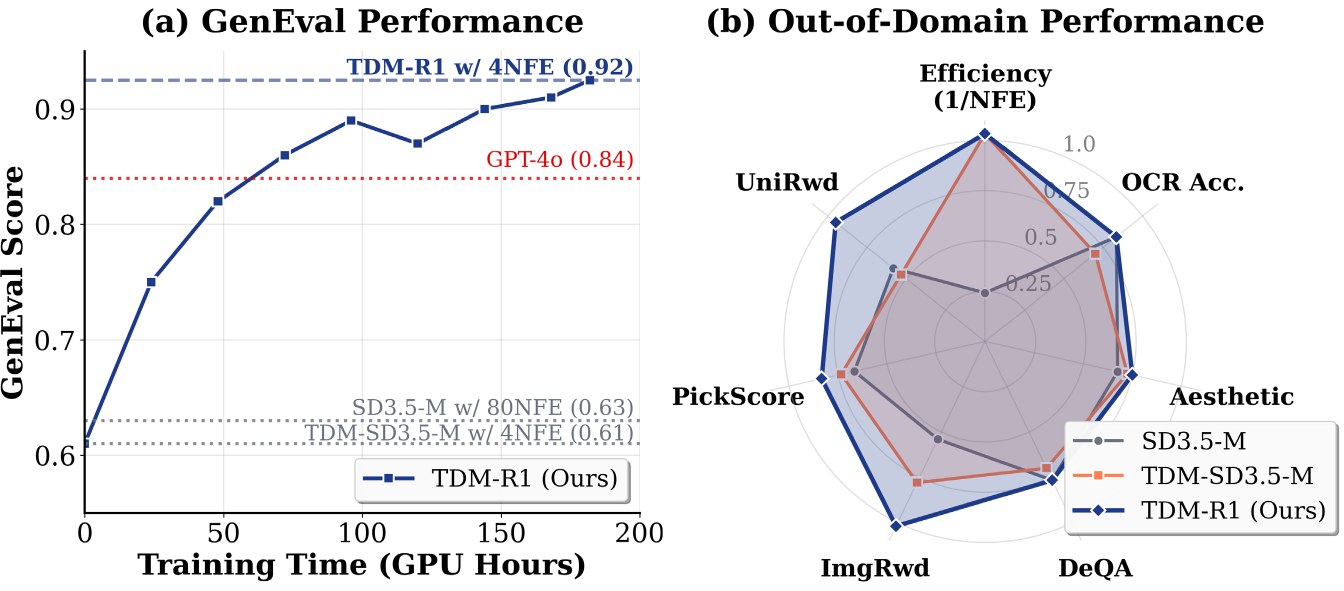}
    \vspace{-2mm}
    \caption{\textbf{TDM-R1} rapidly boosts GenEval score of few-step TDM, notably outperforming its many-step base model and GPT-4o. This is achieved without sacrificing out-of-domain metrics.}
    \label{fig:teaser}
    \vspace{-4mm}
\end{figure}

\begin{figure*}[t]
    \centering
    \includegraphics[width=1\linewidth]{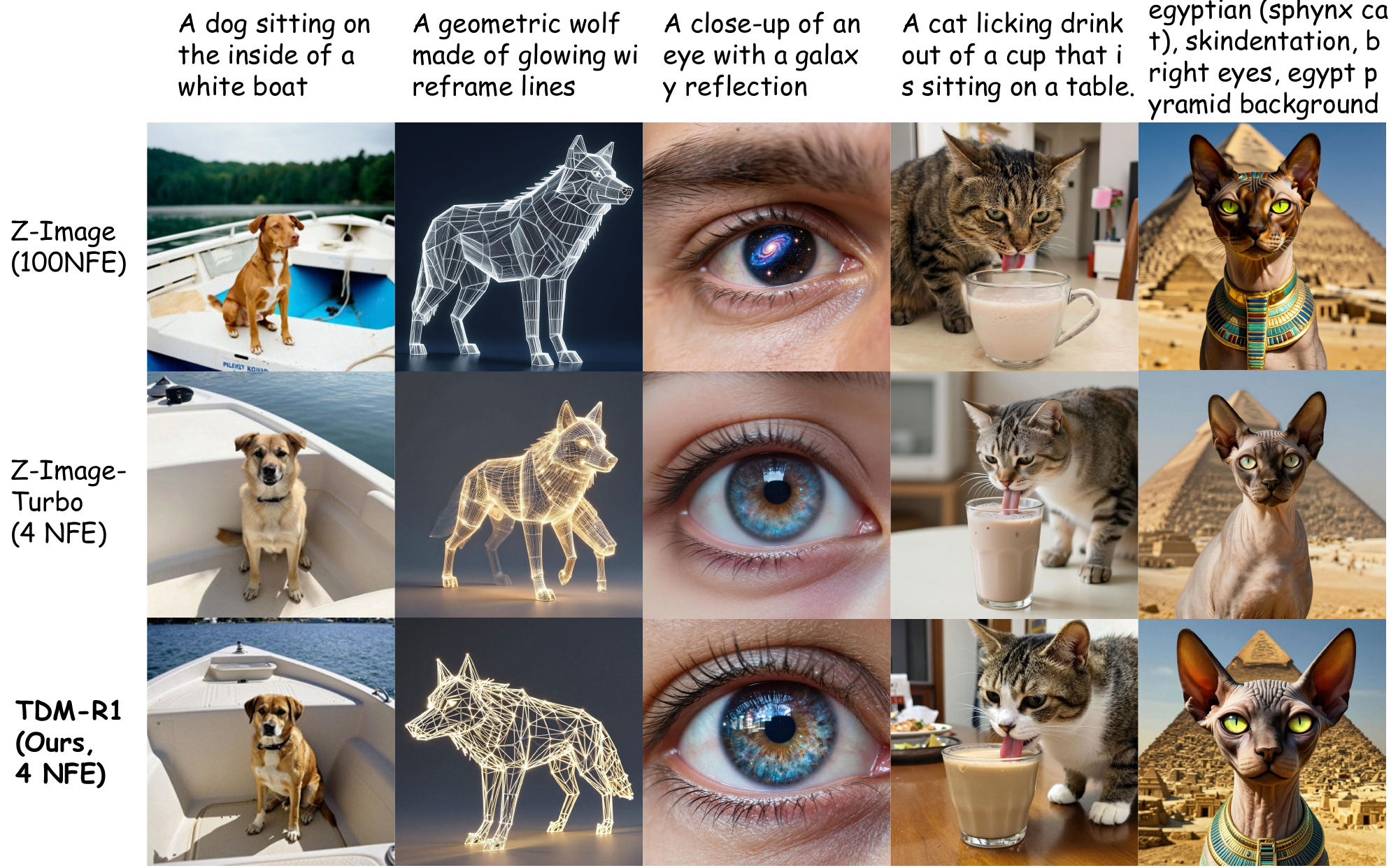}
    \vspace{-5mm}
    \caption{Qualitative comparisons of TDM-R1 against competing methods.}
    \label{fig:visual_compare_z}
    \vspace{-4mm}
\end{figure*}

Though existing papers have explored RL methods for few-step generative models \citep{luo2024diff,ren2024hyper,luo2025reward}, their algorithms rely on a narrow assumption: 
\begin{compactitem}
    \item \textit{\textbf{they require reward signals to be differentiable, such that the outputs of the few-step generative models are able to backpropagate through the rewards.}}
\end{compactitem}

Such a narrow requirement clearly excludes a vast array of essentially non-differentiable reward signals, such as human binary preferences in real-world cases, discrete object countsat, and correctness of text-rendering via OCR models. On the other side, as pointed out by representative processes in LLM, e.g., Deepseek-R1 \citep{deepseek_r1}, rapid successes of RL in large language models clearly indicate the importance of generic non-differential rewards in unlocking models' hidden potentials.

To solve the key problem of \textit{using non-differentiable rewards}, we introduce \textbf{TDM-R1}, a novel reinforcement learning paradigm designed to reinforce these few-step models using \textit{free-form non-differentiable reward feedback} without requiring additional ground-truth image data. \textbf{TDM-R1} is built upon Trajectory Distribution Matching (TDM)~\cite{luo2025learning}, a typical few-step generative model. The core idea of TDM-R1 is to utilize the deterministic trajectories of TDM to assign rewards to intermediate denoising steps via an efficient and unbiased reward estimate for every sample along the trajectory. We also propose a diffusion-parameterized dynamic surrogate reward trained using pairwise groups to provide stable RL supervision at each step. 

By decoupling the learning process into surrogate reward learning and generator optimization, TDM-R1 enables few-step models to achieve performance levels that even surpass many-step counterparts. In Section \ref{sec:exp}, we conduct extensive experiments across compositional image generation, ranging from visual text rendering and human preference alignment to validate the superiority of TDM-R1 on both in-domain and out-of-domain metrics. \textit{\textbf{Remarkably, our method enables models to outperform expensive 80-NFE base models using only 4 NFE}}. Most notably, on the rigorous GenEval benchmark~\cite{ghosh2023geneval}, TDM-R1 boosts performance from 61\% to 92\%—significantly surpassing the 80-NFE base model (63\%) and the commercial state-of-the-art GPT-4o (84\%). Beyond SD3.5-M, TDM-R1 also scales to the powerful 6B-parameter Z-Image model~\cite{zimage}, surpassing both its 100-NFE and few-step variants across in-domain and out-of-domain metrics with only 4 NFEs.

\vspace{-2mm}
\section{Preliminaries}
\vspace{-1mm}

\spara{Diffusion Models} 
Diffusion Models (DMs)\citep{sohl2015deep, ho2020denoising} establish a forward diffusion process that corrupts input sample $\rvx_0$ with Gaussian noise over $T$ discrete timesteps. 
This forward process is defined by: $q(\rvx_t|\rvx) \triangleq \mathcal{N}(\rvx_t; \alpha_t\rvx, \sigma_t^2\textbf{I})$, where $\alpha_t$ and $\sigma_t$ are hyperparameters regarding the noise schedule. 
At any given timestep, noisy samples are generated as $\rvx_t = \alpha_t\rvx + \sigma_t \epsilon$, with $\epsilon \sim \mathcal{N}(\mathbf{0},\mathbf{I})$. The learned reverse diffusion process is formulated as $p_{\psi}(\rvx_{t-1}|\rvx_t) \triangleq \mathcal{N}(\rvx_{t-1}; \mu_\psi(\rvx_t, t), \eta_t^2\textbf{I})$. 
Training proceeds by optimizing the neural network $f_\psi$ through the denoising objective 
\begin{equation}
\label{eq:dsm}
\E_{\rvx,\epsilon,t} \lambda_t || f_\psi(\rvx_t,t) - \rvx||_2^2.
\end{equation}
A well-trained diffusion model can estimate the score via:  $\nabla_{\rvx_t} \log p_t(\rvx_t)\approx s_\psi(\rvx_t) = -\frac{\rvx_t - \alpha_t f_\psi(\rvx_t,t)}{\sigma_t^2}$. 

\spara{Trajectory Distribution Matching (TDM)} TDM~\cite{luo2025learning} aligns the student trajectory and teacher trajectory at the distributional level. The learning is performed by minimizing the integral reverse KL divergence~\cite{luo2023diff,wang2023prolificdreamer} at each step of the $K$-step student trajectory as follows:
\begin{equation}
\begin{aligned}
\label{eq:tdm_grad}
    \nabla_\theta L(\theta) & = \E_{k,t\geq t_k} \lambda_t [ \nabla_{\rvx_{t}}\mathrm{KL}(p_{\theta,k}(\rvx_{t})||p_\psi(\rvx_t))] \frac{\partial \rvx_{t_k}}{\partial \theta} \\
    & \approx \E_{k,t\geq t_k,q(\rvx_t|\rvx_{t_k})} \lambda_t [ s_{\text{fake}}(\rvx_t) - s_\psi(\rvx_t) ] \frac{\partial \rvx_{t_k}}{\partial \theta},
\end{aligned}
\end{equation}
where $t_k=\frac{T}{K}k$ by default, $T$ is the terminal timestep, $p_{\theta,k}(\rvx_{t})\triangleq \int q(\rvx_t|\rvx_{t_k})p_\theta(\rvx_{t_k})d\rvx_{t_k}$, $p_\theta(\rvx_{t_k})$ is the distribution of student trajectory at timestep $t_k$, the $p_\psi(\rvx_t)$ is the pre-trained teacher DM, and a online fake score $s_{\text{fake}}(\rvx_t)$ is employed to approximate the score of student distribution.

\spara{Reward Modeling and RLHF}
Consider ranked pairs derived from a condition $\rvc$ denoted $\rvx^w_0 \succ \rvx^l_0 | \rvc$, where $\rvx^w_0$ represents the preferred sample and $\rvx^l_0$ represents the less preferred sample. 
The Bradley-Terry (BT) model characterizes the pairwise preferences as:
\begin{equation}\label{eq:BT}
p_\text{BT}(\rvx^w_0 \succ \rvx^l_0 | \rvc)=\sigma(r(\rvc,\rvx^w_0)-r(\rvc,\rvx^l_0)),
\end{equation}
where $\sigma(\cdot)$ is the sigmoid function. A parameterized reward network $r_\phi$ can then be learned through maximum likelihood estimation: $\min_\phi -\log p_\text{BT}(\rvx^w_0 \succ \rvx^l_0 | \rvc)$.

The target of RLHF is to optimize a conditional generator $p_\theta(\rvx_0|\rvc)$ such that it maximizes an certain reward $r_\phi(\rvc, \rvx_0)$ while remaining close to a reference distribution $p_{\text{ref}}$ via:
\begin{equation}
\max_{p_\theta} \bbE_{\rvc,\rvx_0\sim p_\theta(\rvx_0|\rvc)} \left[r(\rvc,\rvx_0)\right] - \beta \mathrm{KL}\left[p_\theta(\rvx_0|\rvc)|p_{\text{ref}}(\rvx_0|\rvc)\right].
\label{eq:objective}
\end{equation}
Here, $\beta$ controls the strength of the regularization.

\begin{figure}[!t]
    \centering
    \includegraphics[width=1\linewidth]{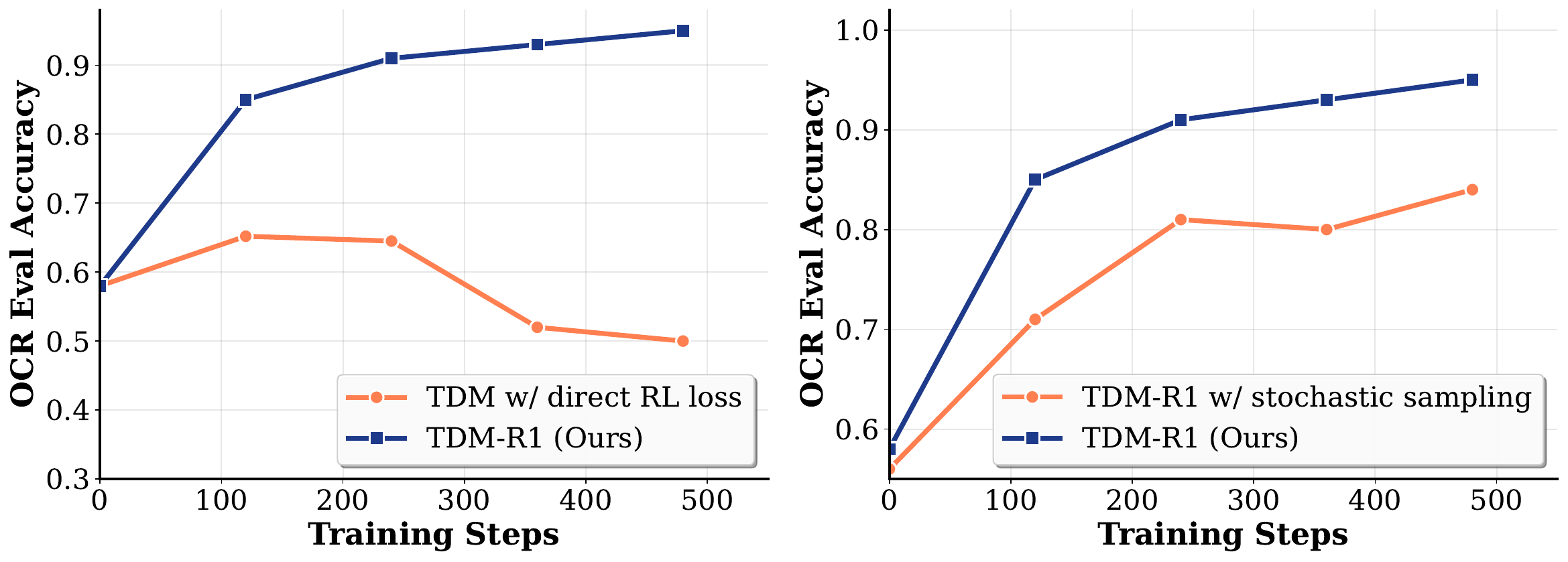}
    \vspace{-6mm}
    \caption{Compare the training performance and speed of TDM-R1 and potential baselines.}
    \label{fig:compare_baseline}
\end{figure}

\vspace{-3mm}
\section{Method}
\vspace{-2mm}

\spara{Problem Setup} 
Let $p_{\theta}$ denote a pre-trained few-step diffusion model parameterized by $\theta$. We assume access to a dataset of conditions $\mathcal{D}_c = \{\rvc_i\}_{i=1}^N$ and a reward function $r(\cdot,\cdot): \mathcal{X} \times \mathcal{C} \rightarrow \mathbb{R}$ that measures the quality of generated samples $\rvx \in \mathcal{X}$ with respect to a given condition $\rvc \in \mathcal{C}$. \textit{Our goal is to reinforce the few-step diffusion model $p_\theta$ according to the non-differentiable reward signal.} We adopt an online RL framework: at each training iteration, the current model $p_{\theta}$ produces a group of samples conditioned on $c \in \mathcal{D}_c$ through $K$-step sampling. The samples generated along each trajectory are collected into datasets
$\mathcal{D}_k = \{(\mathcal{G}_i = \{\rvx_{t_k,j}\}_{j=1}^G, \rvc_i) \mid \rvx_{t_k,j} \sim p_{\theta}(\cdot,t_k|\rvc_i)\}$,
which are subsequently scored by the reward function to obtain training signals.

\spara{Overview} Our method addresses reinforcement learning of few-step diffusion models through three key components. First, in \cref{sec:deterministic}, we establish that deterministic sampling trajectories—as employed by trajectory distribution matching (TDM)—enable accurate reward estimation for intermediate steps (\cref{eq:noisy_reward}). Second, in \cref{sec:surrogate}, we address the incompatibility between standard diffusion RL methods and few-step generation by introducing a Surrogate Reward. This module learns a fine-grained, differentiable reward for each step along the trajectory via group-based preference optimization. Third, in \cref{sec:generator}, we formulate the learning objective for the few-step generator.

\vspace{-1mm}
\subsection{Accurate Intermediate Reward Estimation via Deterministic Trajectories}
\label{sec:deterministic}

In order to leverage reward signals to reinforce a well-trained few-step model. An obvious challenge is that reward signals are generally defined over clean images, while few-step models perform inference by progressively denoising from noise to images, making it difficult to assign rewards to intermediate steps. 
Previous works often directly use the reward at the endpoint of the sampling trajectory as the reward for the entire diffusion path~\cite{flowgrpo,dancegrpo}, which introduces bias into the rewards assigned to intermediate steps. 
Fortunately, prior works~\cite{luo2025adding} have discovered that a conditional probability over noisy images $p(\rvc|\rvx_t)$ can be obtained as follows:
\begin{equation}
\label{eq:noisy_condp}
p(\rvc|\rvx_t) = \int p(\rvc|\rvx) p(\rvx|\rvx_t) dx.
\end{equation}
If we normalize $r(\rvx,\rvc)$ to $[0,1]$ and interpret this value as the probability that image $\rvx$ is a good sample given condition $\rvc$, i.e., $p(\text{"$\rvx$ is a good sample"} | \rvx, \rvc) = r(\rvx,\rvc)$, we can model the reward for the noisy image $\rvx_t$ by:
\begin{equation}
\label{eq:noisy_reward_gt}
r(\rvx_t,\rvc) = \int r(\rvx,\rvc) p(\rvx|\rvx_t) dx = \mathbb{E}_{p(\rvx|\rvx_t)} r(\rvx).
\end{equation}
This \cref{eq:noisy_reward_gt} actually justifies the reasonableness of defining the reward for the entire diffusion path using the reward of $x_0$, which is a single-sample estimate of the \cref{eq:noisy_reward_gt}. The variance of this estimate depends on the variance of $p(\rvx|\rvx_t)$.

In other words, if $p(\rvx|\rvx_t)$ is a deterministic Dirac distribution—i.e., for diffusion, the path from $\rvx_t$ to $x_0$ is obtained via ODE sampling -- then we can obtain an unbiased estimate of the reward for intermediate samples along the sampling path, which will significantly reduce the variance of reward estimation. This also explains why Mix-GRPO achieves superior performance by converting some SDE steps into ODE steps in RL's rollout~\cite{mixgrpo} — it reduces the variance of estimating rewards for intermediate samples along the trajectory.

Through the above analysis, we propose to conduct reinforcement learning based on TDM—a state-of-the-art few-step diffusion model that performs sampling through deterministic trajectories, allowing us to obtain accurate reward estimates for every intermediate sample along the trajectory, thereby achieving superior reinforcement post-training. Compared with the stochastic trajectories, the deterministic trajectories enable more accurate reward estimation for intermidate step, leading to faster convergence and better ultimate performance (\cref{fig:compare_baseline}).

\begin{figure}
    \centering
    \includegraphics[width=1\linewidth]{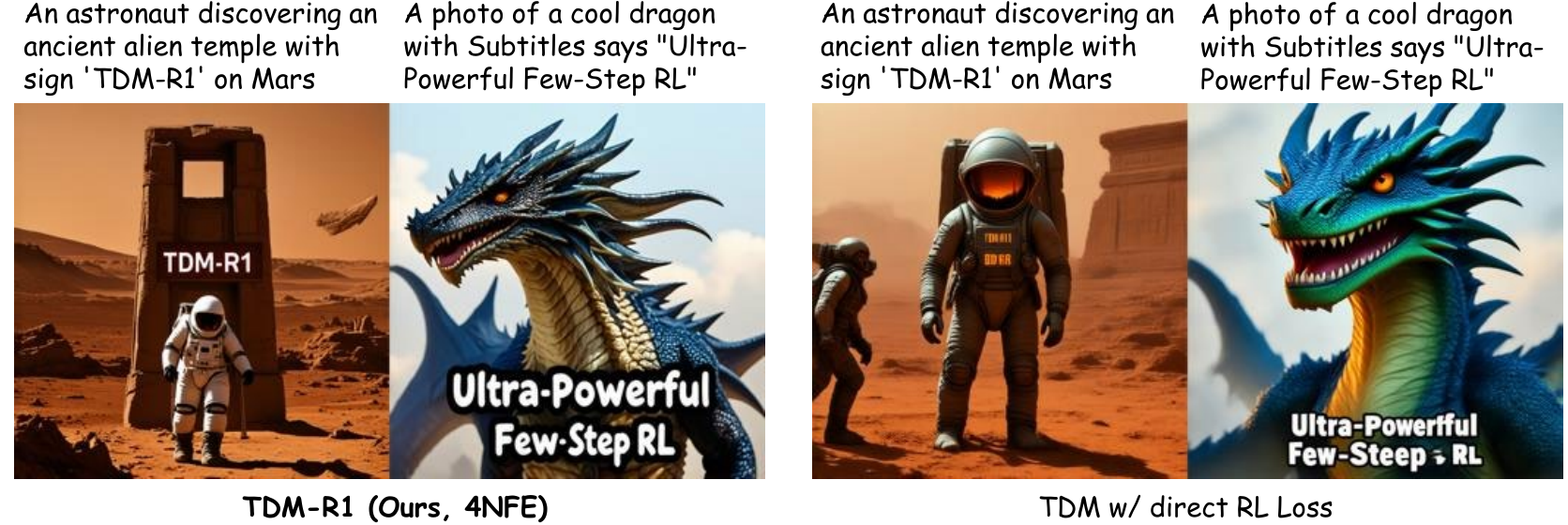}
    \vspace{-4mm}
    \caption{Compare TDM-R1 with the direct combination of TDM and RL loss.}
    \label{fig:compare_niave}
    \vspace{-4mm}
\end{figure}

\subsection{Surrogate Reward Learning} 
\label{sec:surrogate}
Given non-differentiable reward signals and a well-trained few-step diffusion model, a natural idea is to directly apply RL methods designed for standard diffusion models to few-step diffusion models. 
While this approach can leverage reward signals to some extent, it fails to sufficiently improve performance or generate high-quality images—outputs tend to be blurry, as shown in \cref{fig:compare_baseline,fig:compare_niave}.
This is because RL methods for standard diffusion models that handle non-differentiable rewards can typically be reformulated as, or are equivalent to, weighted combinations of denoising losses~\cite{wallace2024diffusion,rl_nft,rl_awm}.
Due to the inherent characteristics of denoising losses, they tend to produce blurry results with few sampling steps, making them incompatible with few-step diffusion models~\cite{luo2025adding}.

To more effectively leverage non-differentiable reward signals, we propose learning a Surrogate Reward for reinforcement and using this Surrogate Reward to guide model learning. Inspired by the success of prior work~\cite{wallace2024diffusion,dgpo}, which have demonstrated the effectiveness of parameterizing reward models using the diffusion models, we adopt a similar approach. Specifically, they derive from the RLHF objective that the reward for a clean image $\rvx_0$ can be approximately parameterized as:
\begin{equation*}
\begin{aligned}
\Tilde{r}_\phi(\rvx_0, \rvc) \approx \beta \E_{q(\rvx_{1:T}|\rvx_0)} \log \frac{p_\theta(\rvx_{0:T}|\rvc)}{p_{\text{ref}}(\rvx_{0:T}|\rvc)} + \beta\log Z(\rvc).
\end{aligned}
\end{equation*}
Through a similar derivation, we can parameterize the surrogate reward for a noisy sample $\rvx_{t_k}$ as follows:
\begin{equation}
\label{eq:noisy_reward}
    \Tilde{r}_\phi(\rvx_{t_k}, \rvc) \approx \beta \E_{q(\rvx_{t_{k+1}:T}|\rvx_{t_k})} \log \frac{p_\phi(\rvx_{t_k:T}|\rvc)}{p_{\text{ref}}(\rvx_{t_k:T}|\rvc)} + \beta\log Z(\rvc).
\end{equation}
One could learn this Surrogate Reward through preference relations between pairwise positive and negative samples. 
However, this approach cannot leverage fine-grained rewards and relative preference relations within groups, limiting the effectiveness of reward learning. 
In contrast, group-based methods have achieved widespread success in both the language models and image generation~\cite{deepseek_r1,flowgrpo}. 

Therefore, we propose directly learning preference relations between pairwise positive and negative groups of noisy samples using the Bradley-Terry (BT) model. Following prior work~\cite{dgpo}, we formulate the learning objective as follows:
\begin{equation}
\label{eq:logp_bt}
\min -\log p(\mathcal{G}_k^+ \succ \mathcal{G}_k^-|\rvc) = -\log \frac{1}{1+\exp( R(\mathcal{G}_k^-) - R(\mathcal{G}_k^+))},
\end{equation}
where $\mathcal{G}_k^+ \cup \mathcal{G}_k^- = \mathcal{G}_k$, and the group-level reward $ R(\mathcal{G})$ is defined as the weighted sum of rewards for each sample within the group, i.e., $R(\mathcal{G}_k) = \sum_{\rvx \in \mathcal{G}_k} w(\rvx_{t_k}) \Tilde{r}_\phi(\rvx_{t_k}, \rvc)$. 

The weight $\omega(\rvx_{t_k})$ controls the importance level of each noisy sample within the group, which is set to the absolute normalized value of the advantage within the group, i.e., $\omega(\rvx_{t_k}^i) = |A(\rvx_{t_k}^i)| = |\frac{r_i - \text{mean}(\{r_j\}_{j=1}^G)}{\text{std}(\{r_j\}_{j=1}^G)}|$. This design assigns larger weights to samples that are either significantly better or worse within the group, providing fine-grained learning signals. 

The partition into positive and negative groups is based on the advantage $A(\rvx_{t_k}^i)$, i.e., 
\begin{equation*}
    \mathcal{G}_k^+ = \{\rvx_{t_k}^i : A(\rvx_{t_k}^i) > 0\}, \  \mathcal{G}_k^- = \{\rvx_{t_k}^i : A(\rvx_{t_k}^i) \leq 0\}.
\end{equation*}
Through some calculation, a tractable upper-bound of loss in \cref{eq:logp_bt} can be rewritten in the following form:
\begin{equation}
\label{eq:kl_dgpo}
\begin{aligned}
    L(\theta) & \leq \E_{(\mathcal{G}_k^+, \mathcal{G}_k^-) \sim \mathcal{D}_k, k} 
    \E_{t\geq t_k,q(\rvx_t|\rvx_{t_k})}
    \log\sigma(  -\beta (T-t_k) \{ \\
    & \sum_{\rvx_{t_k} \in \mathcal{G}_k^+} w(\rvx_{t_k}) [
    \Delta\kl_\phi(\rvx_{t}, \rvx_{t_k}) 
     - \sum_{\rvx_{t_k} \in \mathcal{G}_k^-} w(\rvx_{t_k})  \cdot [
    \Delta\kl_\phi(\rvx_{t}, \rvx_{t_k})
    ]\} 
    ), 
\end{aligned}
\end{equation}
where
\begin{equation*}
    \Delta\kl_\phi(\rvx_{t}, \rvx_{t_k}) \triangleq \kl(q(\rvx_{t-1}|\rvx_t,\rvx_{t_k})||p_\phi(\rvx_{t-1}|\rvx_{t}))
    - \kl(q(\rvx_{t-1}|\rvx_t,\rvx_{t_k})||p_{\text{ref}}(\rvx_{t-1}|\rvx_{t})).
\end{equation*}
In ~\cref{eq:kl_dgpo}, we omitted the condition $\rvc$ for simplicity without loss of generality. Derivations are deferred in  Appendix~\ref{app:derivation}. Under the Gaussian parameterization~\cite{song2020denoising}, the posterior and model distributions are given by: $q(\rvx_{t-1}|\rvx_t, \rvx_{t_k})\triangleq \mathcal{N}(\alpha_{t|t_k} \rvx_{t_k} + \sqrt{\sigma_{t-1|t_k}^2 - \eta_t^2}\epsilon, \eta_t^2I )$ and $p_{\phi}(\rvx_{t-1}|\rvx_t) \triangleq \mathcal{N}(\alpha_{t-1}\frac{\rvx_t - \sigma_t\epsilon_\phi}{\alpha_t} + \sqrt{\sigma_t^2 - \eta_t^2}\epsilon_\phi, \eta_t^2I )$.

\spara{Dynamic Reference Model} 
A frozen reference model may impose overly strong regularization, thereby hindering the learning of the reward model. 
Prior work has opted to update the reference model with theta at fixed intervals. 
However, this approach may introduce instability: $\theta$ might overfit to noisy signals at a certain step and subsequently be used as the reference model. 
To achieve better dynamic updating of the reference model, we propose to use an EMA (Exponential Moving Average) version of $\phi$ as the reference model. 
This not only relaxes the regularization to facilitate better reward learning, but also mitigates the risk of adopting a ``bad" reference model that has overfitted to noisy signals.

\begin{algorithm}[t]
\caption{TDM-R1}
\label{alg:tdm_r1}
\begin{algorithmic}[1]
\Require Few-step diffusion model $p_\theta$, surrogate reward model $\tilde{r}_\phi$, reward function $r$, group size $G$, number of iterations $N$, number of denoising steps $K$.
\Ensure Reinforced few-step model $p_\theta$.

\For{$n \leftarrow 1$ {\bfseries to} $N$}

\State \textcolor{gray}{\textit{// Sample conditioning and generate group}}
\State Sample conditioning $\mathbf{c} \sim \mathcal{D}_c$
\State Generate group $\mathcal{G} = \{\mathbf{x}_0^{(1)}, \ldots, \mathbf{x}_0^{(G)}\}$ by sampling from $p_{\theta^-}(\cdot \mid \mathbf{c})$ with $K$ steps

\State \textcolor{gray}{\textit{// Access reward feedback}}
\State Compute rewards: $r_i \gets r(\mathbf{c}, \mathbf{x}_0^{(i)})$ for $i = 1, \ldots, G$

\State \textcolor{gray}{\textit{// Update models}}
\State Update surrogate reward $\tilde{r}_\phi$ via \cref{eq:kl_dgpo}
\State Update fake score $s_{\text{fake}}$ via denoising on perturbed generated samples (\cref{eq:dsm}).
\State Update few-step generator $p_\theta$ via \cref{eq:k_rlhf}

\EndFor
\end{algorithmic}
\end{algorithm}

\subsection{Few-Step Generator Learning}
\label{sec:generator}

We formulate the reinforcement learning training objective for the $K$-step generator in a form similar to standard RLHF~\cite{luo2023diff,luo2024diffstar}, consisting of reward maximization with a reverse KL regularization as follows:
\begin{equation}
\label{eq:k_rlhf}
    L(\theta) =  \E_{k,p_{\theta}(\rvx_{t_k})} - \Tilde{r}_{\mathrm{sg}(\phi)}(\rvx_{t_k},\rvc) + \beta_g \mathrm{KL}(p_{\theta,k}(\rvx_t)||p_\psi(\rvx_t)),
\end{equation}
where $p_{\theta,k}(\rvx_t)\triangleq \int q(\rvx_{t}|\rvx_{t_k})p_{\theta}(\rvx_{t_k}) d\rvx_{t_k}$. 
In the objectives described above, Surrogate Reward maximization provides a good learning signal to integrate non-differentiable reward feedback, while the marginal-level reverse KL regularization effectively grounds the generated samples to the base distribution parameterized by the pretrained diffusion model at the distribution level. 
This differs from the KL regularization adopted in standard diffusion RL~\cite{fan2024reinforcement,flowgrpo}, which is essentially an instance-level constraint that requires each point along the trajectory to be consistent with the base model, leading to an unnecessarily difficult constraint.

Put the reward parameterization in \cref{eq:noisy_reward} into \cref{eq:k_rlhf}, we can derive the learning gradient with some calculations as follows:
\begin{equation}
\label{eq:tdm_r1_grad}
\begin{aligned}
    \nabla_\theta L(\theta) & = - \E_{k, t\geq t_k} \E_{p_{\theta}(\rvx_{t_k})} \E_{q(\rvx_t,\rvx_{t-1}|\rvx_{t_k})} \Bigg[ \beta\alpha_{t|t_k}(T-t_k) \nabla_{\rvx_{t}} \log \frac{p_\phi(\rvx_{t-1}|\rvx_{t})}{p_{\text{ref}}(\rvx_{t-1}|\rvx_{t})} \\ 
    & + \beta_g \lambda_t \left( s_{\text{fake}}(\rvx_t) - s_\psi(\rvx_t) \right) \Bigg] \frac{\partial \rvx_{t_{k}}}{\partial \theta},
\end{aligned}
\end{equation}
where we stop the gradient of $\phi$ to save the memory cost, and we found that it does not affect the performance. We defer the derivation to Appendix~\ref{app:derivation}.
Following standard practice~\cite{luo2023diff,luo2025learning,luo2024one,yin2023one,zhou2024score}, we adopt an online fake score $s_{fake}$ for estimating the score of student distribution, which is trained by denoising score matching.

\spara{Remark} 
By joint training of the Few-Step Generator and Surrogate Reward, which establishes a synergistic loop: the Generator progressively produces higher-quality samples via maximize the Surrogate Reward, while the Surrogate Reward adapts to provide increasingly precise guidance by identifying favorable and unfavorable regions at each intermediate step per iteration. This Adaptive Surrogate Reward mechanism implements a GAN-like adversarial framework on self-generated samples, ultimately enabling effective reinforcement post-training for few-step diffusion models with superior generation performance.

In summary, TDM-R1 enables the utilization of large-scale online non-differentiable reward feedback, achieving effective reinforcement post-training for few-step generators. Starting from a pre-trained few-step generator based on Trajectory Distribution Matching (TDM), we alternately optimize the few-step generator to maximize the surrogate reward and minimize the reverse KL divergence, while the surrogate reward is trained through group preference optimization, and the fake score is optimized via the denoising matching objective. See Algorithm~\ref{alg:tdm_r1} for pseudo algorithm.

\begin{table*}[!t]
    \centering
    \caption{{\bf GenEval Result.} We {\bf highlight} the best scores among specific category. Obj.: Object; Attr.: Attribution.}
    \vspace{-4mm}
    \resizebox{\linewidth}{!}{
    \begin{tabular}{l|c|cccccc}
    \toprule
    \textbf{Model} & \textbf{Overall} & \textbf{Single Obj.} & \textbf{Two Obj.} & \textbf{Counting} & \textbf{Colors} & \textbf{Position} & \textbf{Attr. Binding} \\
    \midrule
    \multicolumn{8}{l}{\textbf{\textit{Autoregressive Models:}}} \\
    \midrule
    Show-o~\citep{xie2024show}  & 0.53 & 0.95 & 0.52 & 0.49 & 0.82 & 0.11 & 0.28 \\
    Emu3-Gen~\citep{wang2024emu3}  & 0.54 & 0.98 & 0.71 & 0.34 & 0.81 & 0.17 & 0.21 \\
    JanusFlow~\citep{ma2025janusflow}  & 0.63 & 0.97 & 0.59 & 0.45 & 0.83 & 0.53 & 0.42 \\
    Janus-Pro-7B~\citep{chen2025janus}  & 0.80 & \textbf{0.99} & 0.89 & 0.59 & 0.90 & \textbf{0.79} & \textbf{0.66} \\
    GPT-4o~\citep{hurst2024gpt} & \textbf{0.84} & \textbf{0.99} & \textbf{0.92} & \textbf{0.85} & \textbf{0.92} & 0.75 & 0.61 \\
    \midrule
    \multicolumn{8}{l}{\textbf{\textit{Diffusion Models:}}} \\
    \midrule
    LDM~\citep{rombach2022high}  & 0.37 & 0.92 & 0.29 & 0.23 & 0.70 & 0.02 & 0.05 \\
    SD1.5~\citep{rombach2022high}  & 0.43 & 0.97 & 0.38 & 0.35 & 0.76 & 0.04 & 0.06 \\
    SD2.1~\citep{rombach2022high}  & 0.50 & 0.98 & 0.51 & 0.44 & 0.85 & 0.07 & 0.17 \\
    SD-XL~\citep{podell2023sdxl}  & 0.55 & 0.98 & 0.74 & 0.39 & 0.85 & 0.15 & 0.23 \\
    DALLE-2~\citep{dalle2}  & 0.52 & 0.94 & 0.66 & 0.49 & 0.77 & 0.10 & 0.19 \\
    DALLE-3~\citep{betker2023dalle3}  & 0.67 & 0.96 & 0.87 & 0.47 & 0.83 & 0.43 & 0.45 \\
    FLUX.1 Dev~\citep{flux2024}  & 0.66 & 0.98 & 0.81 & 0.74 & 0.79 & 0.22 & 0.45 \\
    SD3.5-L~\citep{sd3}  & 0.71 & 0.98 & 0.89 & 0.73 & 0.83 & 0.34 & 0.47 \\
    SANA-1.5 4.8B~\citep{xie2025sana} & 0.81 & 0.99 & 0.93 & 0.86 & 0.84 & 0.59 & 0.65 \\
    \midrule
    SD3.5-M~\citep{sd3} & 0.63 & 0.98 & 0.78 & 0.50 & 0.81 & 0.24 & 0.52 \\
    w/ Flow-GRPO~\citep{flowgrpo} & 0.95 & \textbf{1.00} & 0.99 & 0.95 & 0.92 & \textbf{0.99} & 0.86  \\
    w/ DGPO~\cite{dgpo} & \textbf{0.97} & \textbf{1.00} & \textbf{0.99} & \textbf{0.97} & \textbf{0.95} & \textbf{0.99} & \textbf{0.91}  \\
    \midrule
    \multicolumn{8}{l}{\textbf{\textit{Few-Step Diffusion Models:}}} \\
    \midrule
    SD3.5-L-Turbo (4NFE)~\citep{sd3} & 0.70 & 0.94 & 0.84 & 0.55 & 0.79 & 0.58 & 0.56 \\
    TDM-SD3.5-M (4NFE)~\cite{luo2025learning} & 0.61 & 0.99 & 0.77 & 0.49 & 0.79 & 0.23 & 0.44 \\
    \textbf{SD3.5-M  w/ TDM-R1 (Ours, 4NFE)} & \textbf{0.92} & \textbf{1.00} & \textbf{0.96} & \textbf{0.88} & \textbf{0.85} & \textbf{0.93} & \textbf{0.91}\\
    \bottomrule
    \end{tabular}
    }
    \label{tab:geneval}
    \vspace{-4mm}
\end{table*}

\begin{table*}[!t]
    \centering
    \renewcommand{\arraystretch}{1.1}
    \caption{\textbf{Performance on Compositional Image Generation and Visual Text Rendering Benchmarks}. ImgRwd: ImageReward; UniRwd: UnifiedReward. We \textcolor{blue}{highlight} the metric adopted for the training signal. $^\dagger$ Results are taken from the original paper.}
    \vspace{-2mm}
    \resizebox{\linewidth}{!}{
        \begin{tabular}{lccccccccc}
            \toprule
            \multirow{2}{*}{\textbf{Model}} & \multirow{2}{*}{\textbf{NFE}} & \multicolumn{2}{c}{\textbf{Verifiable Metric}} & \multicolumn{2}{c}{\textbf{Image Quality}} & \multicolumn{3}{c}{\textbf{Preference Score}}    \\ \cmidrule(lr){3-4} \cmidrule(lr){5-6} \cmidrule(l){7-9} 
                                   & & \textbf{GenEval}  & \textbf{OCR Acc.}  & \textbf{Aesthetic}          & \textbf{DeQA}         & \textbf{ImgRwd} & \textbf{PickScore} & \textbf{UniRwd} \\ \midrule
SD3.5-M~\cite{sd3} & 80 & 0.63 & 0.59 & 5.39 & 4.07 & 0.87 & 22.34 & 3.33 \\
\midrule
Flow-GRPO$^\dagger$~\cite{flowgrpo}  & 80 & \textcolor{blue}{0.95} & --- & 5.25 & 4.01 & 1.03 & 22.37 & 3.51 \\
DGPO$^\dagger$~\cite{dgpo} & 80 & \textcolor{blue}{0.97} & --- & 5.31 & 4.03 & 1.08 & 22.41 & 3.60 \\
\midrule
Flow-GRPO$^\dagger$~\cite{flowgrpo}  & 80 & --- & \textcolor{blue}{0.92} & 5.32 & 4.06 & 0.95 & 22.44 & 3.42 \\
DGPO$^\dagger$~\cite{dgpo}  & 80 & --- & \textcolor{blue}{0.96} & 5.37 & 4.09 & 1.02 & 22.52 & 3.48 \\
\midrule 
\midrule
TDM-SD3.5-M~\cite{luo2025learning} & 4 & 0.61 & 0.55 & 5.41 &  4.05 & 0.99 & 22.36 & 3.30 \\
\midrule
\multicolumn{9}{l}{\textbf{\textit{Compositional Image Generation:}}} \\
\midrule
$p_\phi$ in Surrogate Reward & 80 & \textcolor{blue}{0.89} & 0.52 & 5.35 & 4.01 & 0.94 & 22.31 & 3.48\\
\textbf{TDM-R1 (Ours)} & 4 & \textcolor{blue}{0.92} & 0.59 & 5.42 &  4.07 & 1.11 & 22.39 & 3.55\\
\midrule
\multicolumn{9}{l}{\textbf{\textit{Visual Text Rendering:}}} \\
\midrule
$p_\phi$ in Surrogate Reward & 80 & 0.64 & \textcolor{blue}{0.92} & 5.41 & 4.05 &   0.98 & 22.37 & 3.43\\
\textbf{TDM-R1 (Ours)} & 4 & 0.67 & \textcolor{blue}{0.95} & 5.45 & 4.11 & 1.09 & 22.62 & 3.51 \\
\bottomrule
\end{tabular}%
}
\vspace{-2mm}
\label{tab:main_res}
\end{table*}

\begin{figure*}[t]
    \centering
    \includegraphics[width=1\linewidth]{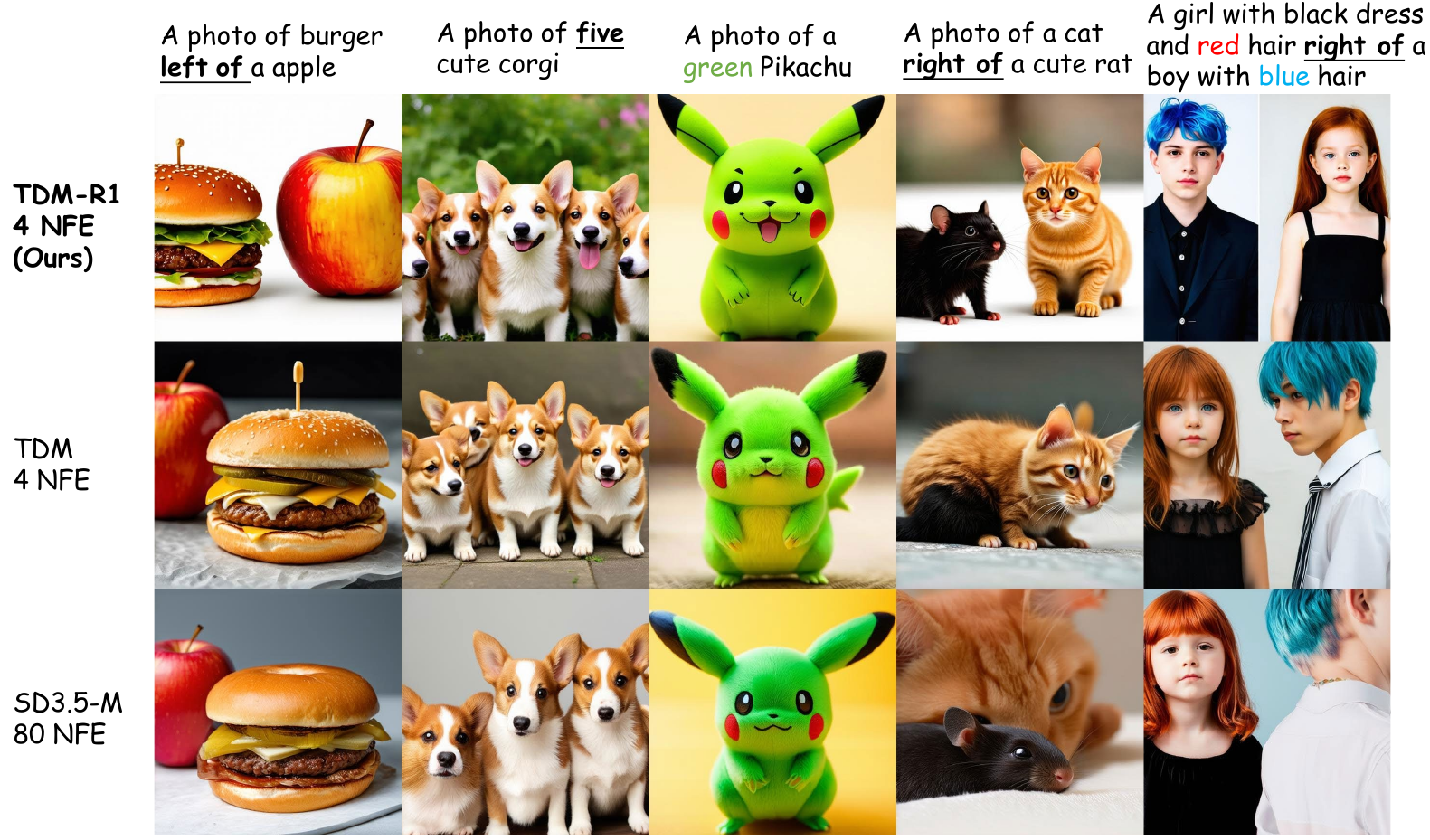}
    \vspace{-5mm}
    \caption{Qualitative comparisons of TDM-R1 with GenEval signals against competing methods. It can be seen that our proposed TDM-R1 can accurately follow the instructions, while keeping a strong visual quality. The same initial noise is used to generate all images.}
    \label{fig:visual_compare}
    \vspace{-2mm}
\end{figure*}

\begin{table*}[!t]
    \centering
    \renewcommand{\arraystretch}{1.1}
    \caption{\textbf{Performance on reinforcing Z-Image}. ImgRwd: ImageReward; UniRwd: UnifiedReward. We \textcolor{blue}{highlight} the metric adopted for the training signal.}
    \vspace{-2mm}
    \resizebox{\linewidth}{!}{
        \begin{tabular}{lccccccccc}
            \toprule
            \multirow{2}{*}{\textbf{Model}} & \multirow{2}{*}{\textbf{NFE}} & \multicolumn{2}{c}{\textbf{Verifiable Metric}} & \multicolumn{5}{c}{\textbf{Model-based Metric}}   \\ \cmidrule(lr){3-4} \cmidrule(lr){5-9} 
                                   & & \textbf{GenEval}  & \textbf{OCR Acc.}  & \textbf{HPSv3}         & \textbf{Aesthetic}          & \textbf{ImgRwd} & \textbf{PickScore} & \textbf{UniRwd} \\ \midrule
Z-Image & 100 & 0.66 & 0.74 & 7.32 & 5.35 & 0.62 & 19.98 & 3.64 \\
Z-Image-Turbo (DMDR) & \textbf{4} & 0.73 & 0.78 & 9.13 & 5.40 & 0.78 & 20.28 & 3.70 \\
\midrule
\textbf{TDM-R1-Zimage (Ours)} & \textbf{4} & \textbf{0.77} & \textbf{0.79} & \textbf{\textcolor{blue}{9.90}} & \textbf{5.49} & \textbf{0.94} & \textbf{20.45} & \textbf{3.75} \\
\bottomrule
\end{tabular}%
}
\vspace{-4mm}
\label{tab:z_res}
\end{table*}

\vspace{-2mm}
\section{Experiments}\label{sec:exp}
\vspace{-1mm}

This section provides a comprehensive evaluation of TDM-R1. We primarily benchmark performance on two verifiable tasks with non-differentiable metrics: compositional image generation and visual text rendering (\cref{tab:geneval,tab:main_res}). Qualitative comparisons are shown in \cref{fig:visual_compare}, and we demonstrate the method's effectiveness in aligning with human preferences (\cref{fig:align_human}). Additionally, we ablate key components of our approach (\cref{fig:compare_baseline,fig:variants_part2}).

\subsection{Experimental Setup}

\spara{Evaluation Tasks} We begin by pre-training the few-step TDM on SD3.5-M~\cite{sd3,luo2025learning}, followed by evaluating our proposed TDM-R1's effectiveness in reinforcing the TDM-SD3.5-M across two valuable tasks with verifiable metrics: 1) \textit{Compositional Image Generation}: assessed using GenEval~\citep{ghosh2023geneval}, which encompasses six demanding compositional generation scenarios, including object counting, spatial relationships, and attribute binding; 
2) \textit{Visual Text Rendering}: measures the model's capability to accurately synthesize text within generated images~\citep{gong2025seedream}. 

\spara{Out-of-Domain Evaluation Metrics} 
To ensure fair assessment and mitigate reward hacking---a phenomenon where models overfit to training reward signals at the expense of actual image quality---we utilize five independent image quality metrics excluded from training as out-of-domain evaluations: Aesthetic Score~\citep{schuhmann2022laion}, DeQA~\citep{deqa}, ImageReward~\citep{xu2023imagereward}, PickScore~\citep{kirstain2023pick}, and UnifiedReward~\citep{wang2025unified}. These metrics are computed on DrawBench~\citep{drawbench}, a comprehensive benchmark comprising diverse prompts.

\subsection{Main Results}

\spara{Quantitative Results} \cref{tab:geneval} shows that our proposed TDM-R1 achieves competitive performance on GenEval, notably surpassing prior commercial closed-source SOTA GPT-4o and matches the performance of prior SOTA standard diffusion RL. \textit{This promising performance is achieved while also showing improvement across various out-of-domain metrics (such as AeS, DeQA, PickScore, and Image Reward), as indicated by \cref{tab:main_res}.} It is worth noting that, although previous SOTA of standard diffusion RL (e.g., Flow-GRPO and DGPO) achieved higher GenEval scores, they exhibited noticeable degradation in metrics measuring image quality. In contrast, our TDM-R1 even achieves higher image quality metrics compared to both the many-step base model and the few-step base model.

Beyond compositional image generation, \cref{tab:main_res} provides detailed evaluation results on visual text rendering, where DGPO similarly demonstrates significant improvements in both target optimization metrics and out-of-domain metrics. \textit{Notably, we found that training on a verifiable metric (GenEval score or OCR score) can synergistically improve a completely different verifiable metric.} This is a surprising and encouraging signal, suggesting that we may be able to enhance the broad instruction-following capabilities of few-step diffusion models through a well-chosen proxy task.

\spara{Qualitative Comparison} 
We present qualitative results from our method and baselines, trained with GenEval's signal, in \cref{fig:visual_compare}. The visualizations demonstrate that TDM-R1 follows instructions more accurately than both the few-step baseline and the 80-NFE base model, while preserving high generation quality. See Appendix~\ref{app:add_samples} for additional visual samples.

\begin{figure}[!t]
    \centering
    \includegraphics[width=1\linewidth]{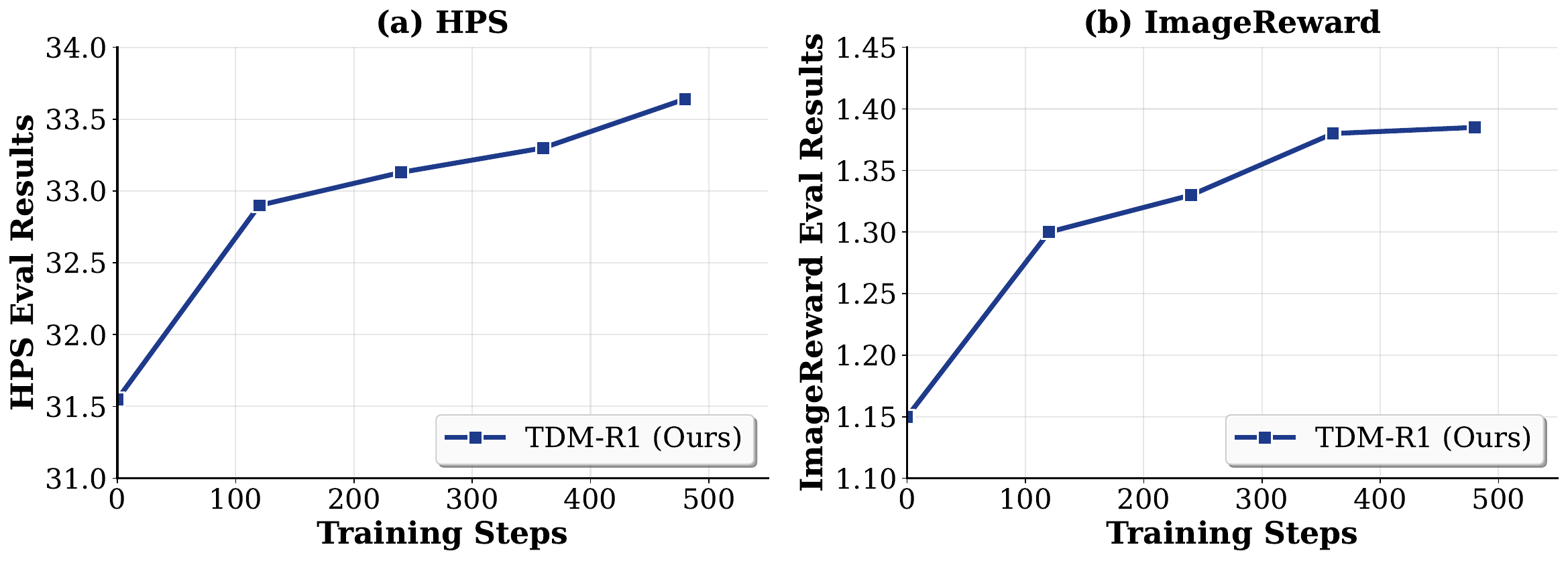}
    \vspace{-6mm}
    \caption{TDM-R1 Performance in human preference alignment.}
    \label{fig:align_human}
    \vspace{-4mm}
\end{figure}

\spara{Human Preference Alignment} We also evaluate TDM-R1 on Human Preference Alignment. In this task, we employ ImageReward~\citep{xu2023imagereward} and HPS~\cite{wu2023human} as the reward signal separately. Although these metrics are differentiable, we do not use their gradient. Both HPS and ImageReward are trained on large-scale human preference data, providing a comprehensive assessment of generation quality from a human-centric perspective. As shown in \cref{fig:align_human}, our proposed TDM-R1 can effectively enhance the performance of few-step models with these metrics that evaluate human preference.

\spara{Human Preference Alignment on Large Model} 
We further evaluate TDM-R1 on aligning a larger model, the recent and powerful open-source Z-Image~\cite{zimage} with 6B parameters, using HPSv3 as the reward signal. HPSv3 is a state-of-the-art reward model that serves as a robust metric for wide-spectrum image evaluation. As shown in \cref{tab:z_res}, TDM-R1 effectively enhances the performance of Z-Image across both in-domain and out-of-domain metrics, notably outperforming the many-step Z-Image and the few-step Z-Image-Turbo. We also provide visual comparisons against competing methods in \cref{fig:visual_compare_z}, demonstrating consistent improvement over the baselines

\spara{Comparison with the Surrogate Reward Model} 
Since our Surrogate Reward is parameterized by a diffusion model $p_\phi$, an interesting experimental question arises: how does the performance of the diffusion model $p_\phi$ that parameterizes this reward compare to our few-step generator TDM-R1? As shown in \cref{tab:main_res}, $p_\phi$ achieves substantially better task metrics than the base model. Notably, however, TDM-R1---which requires only 4 NFE for sampling---consistently outperforms $p_\phi$ (which requires 80 NFE) on both in-domain and out-of-domain metrics. 
This result may appear counterintuitive, given that TDM-R1 is trained using signals derived from the very Surrogate Reward parameterized by $p_\phi$. 
Yet this observation aligns with findings from the LLM literature on DPO-like methods: improving a model's performance as a reward signal does not necessarily translate to improved generation performance in inference. 
The key insight is that effective training of TDM-R1 depends not on $p_\phi$'s generation capabilities, but rather on the quality of the reward signal it provides. These results demonstrate that our Surrogate Reward has been effectively tailored as a per-step reward model, enabling superior reinforcement fine-tuning performance for TDM-R1.

\subsection{Ablation Study}

\spara{TDM-R1 v.s. Direct RL Loss Combination} 
A fairly straightforward and natural baseline is to directly combine the distillation loss with an RL loss designed for standard diffusion models to reinforce the few-step generator. 
We adopt DGPO~\cite{dgpo}, a recent state-of-the-art RL method for diffusion models, as the accompanying RL loss and combine it with TDM's distillation loss as a potential baseline, termed TDM w/ direct RL loss. 
As shown in \cref{fig:compare_baseline,fig:compare_niave}, while this baseline achieves modest performance improvements in the early stages (though still far inferior to TDM-R1), it produces blurry images and suffers from performance degradation in later training stages. This decline in image quality stems from a fundamental mismatch: the denoising objective inherent in standard diffusion RL is incompatible with reverse KL divergence minimization in distillation, as the direct denoising behavior enforced by RL loss may be suboptimal for few-step generation.

\begin{figure}[!t]
    \centering
    \includegraphics[width=1\linewidth]{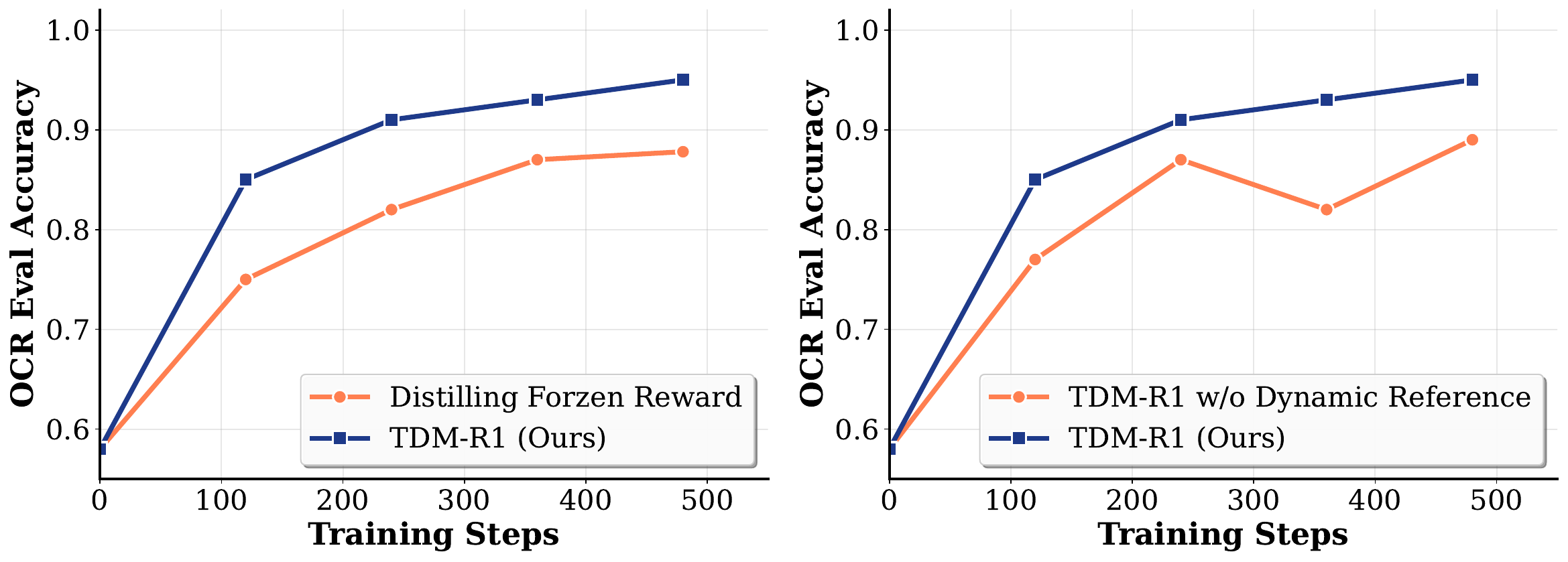}
    \vspace{-6mm}
    \caption{Comparison of visual text alignment across variants.}
    \label{fig:variants_part2}
    \vspace{-4mm}
\end{figure}

\spara{Dynamic Surrogate Reward v.s. Frozen Well-Trained Reward} 
Our proposed TDM-R1 jointly trains the Dynamic Surrogate Reward and the few-step generator throughout the training process. 
We compared this approach against a notable baseline: directly using the reward model trained by DGPO. 
This baseline can be viewed as using an RL-trained diffusion model as a teacher to distill a few-step student. 
As shown in \cref{fig:variants_part2}, the Dynamic Surrogate Reward achieves faster reward growth and superior final performance compared to distillation with a frozen reward. We attribute this improvement to two factors.
First, the Dynamic Surrogate Reward can dynamically identify regions where the few-step student performs well or poorly, enabling more targeted optimization—a capability that the frozen reward lacks. 
Second, a distribution gap exists between the few-step generator's outputs and the data distribution on which the frozen reward was originally trained, which can lead to suboptimal guidance and degraded performance.

\begin{figure}[!t]
    \centering
    \includegraphics[width=0.7\linewidth]{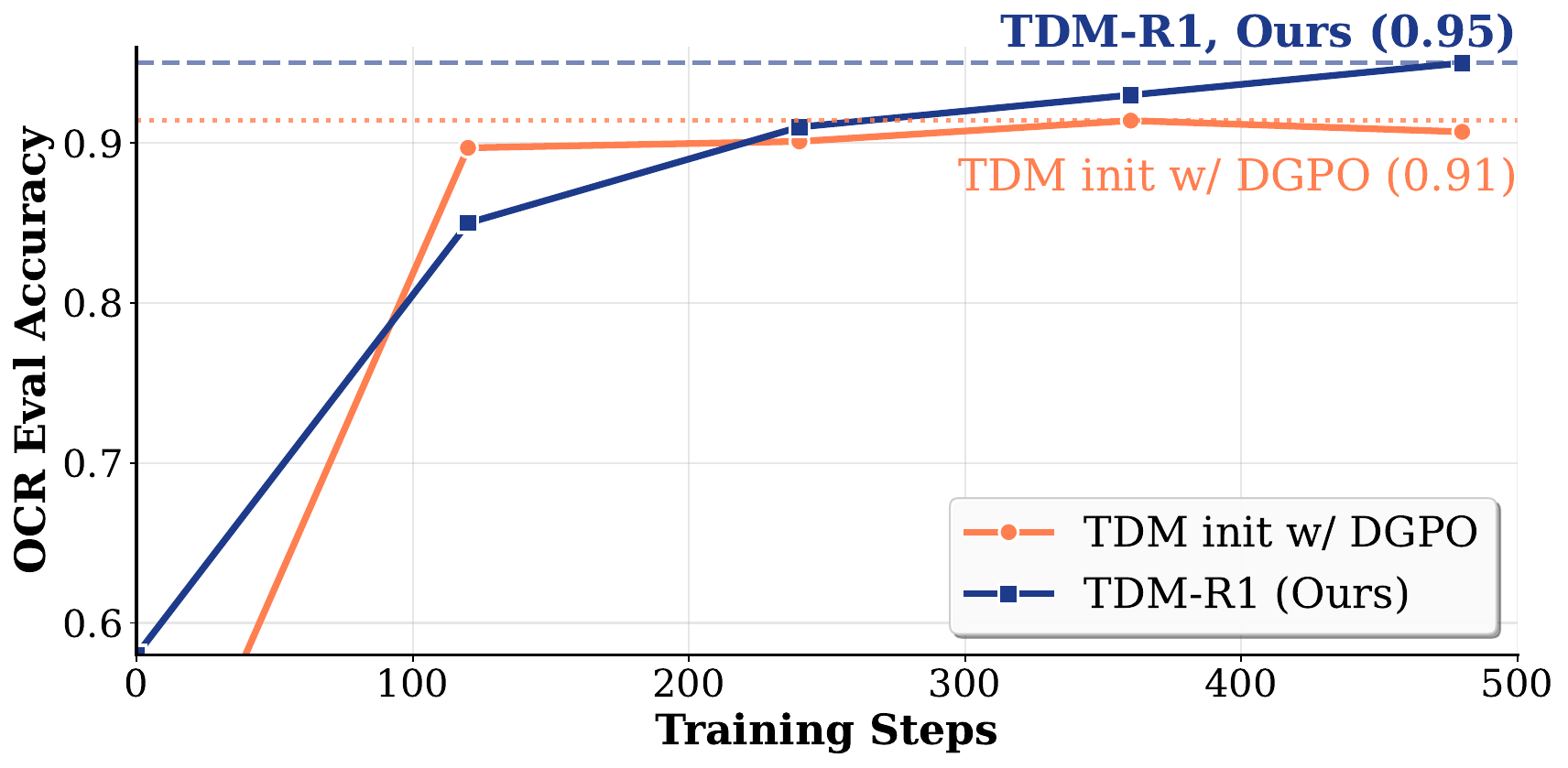}
    \vspace{-3mm}
    \caption{Comparison between TDM-R1 with direct distillation.}
    \label{fig:compare_tdm_dgpo}
\end{figure}

\spara{Comparison with Distilled RL Diffusion} We also compare TDM-R1 against a competitive baseline: initializing with an RL-finetuned diffusion model (i.e., DGPO) followed by distillation via TDM. As shown in \cref{fig:compare_tdm_dgpo}, TDM initialized with DGPO converges quickly in the early stages, but its performance ceiling is constrained by the teacher model, causing it to plateau rapidly. In contrast, TDM-R1 continuously incorporates reward signals throughout training, ultimately achieving superior performance. Although this baseline is beyond our scope --- our goal is to directly post-train few-step diffusion models rather than first training a teacher and then distilling a few-step student, which introduces additional pipeline complexity --- this ablation clearly demonstrates that \textit{TDM-R1 is not only more elegant as a paradigm, directly reinforcing the well-trained few-step student model, but also achieves stronger final performance.}

\spara{Effect of the Deterministic Path} 
A core feature of our method is its ability to leverage deterministic trajectories for step-by-step reward feedback. 
While stochastic trajectories can theoretically achieve the same goal, they introduce greater variance in the feedback signal. 
To evaluate this trade-off, we compared our approach against a variant using stochastic sampling, termed TDM-R1 w/ stochastic sampling. As shown in \cref{fig:compare_baseline}, deterministic trajectories yield faster convergence and superior performance.

\spara{Effect of the Dynamic Reference Model} 
We propose using an EMA of $p_\phi$ as the reference model during surrogate reward learning. This dynamic reference model enables effective optimization. Our experiments confirm the effectiveness of this design: as shown in Figure~\ref{fig:variants_part2}, replacing the dynamic reference model with a static one results in degraded performance and reduced training stability.

\vspace{-1mm}
\section{Related Works}
\vspace{-1mm}

\spara{Few-Step Text-to-Image Diffusion Sampling} Although training-free methods for accelerating diffusion model sampling have progressed substantially~\cite{lu2022dpm, zhao2023unipc, xue2024accelerating, si2024freeu, ma2024surprising, li2025self}, diffusion distillation~\citep{luo2023comprehensive} remains essential for high-quality few-step generation. In general, distilled models perform one or a few more transformations that map random noise to images. The most widely adopted distillation strategies for few-step diffusion sampling fall into two categories: trajectory matching~\citep{song2023consistency,kim2023consistency,songimproved,geng2024consistency,salimans2024multistep} and distribution matching~\citep{luo2023diff,yin2023one,zhou2024score,luo2024one,sauer2023adversarial,xu2024ufogen,yoso,wang2025uni}. More recently, trajectory distribution matching (TDM)~\cite{luo2025learning} has demonstrated strong few-step performance. 

\spara{Text-Image Alignment} 
Among the many efforts to improve text-image alignment~\citep{ho2022classifier,bansal2023universal, ma2023elucidating,liu2024referee,luo2025noise}, reinforcement learning has emerged as a particularly promising approach.  
Recent efforts to improve diffusion models through reinforcement learning have explored three main strategies. The first fine-tunes diffusion on high-quality image-prompt pairs~\citep{dai2023emu,podell2023sdxl}. The second optimizes explicit rewards, either by evaluating the final outputs of multi-step synthesis~\citep{prabhudesai2023aligning,clark2023directly,lee2023aligning,ho2022imagen,luo2025reward,luo2025adding} or through policy gradient methods~\citep{fan2024reinforcement,black2023training,ye2024schedule}. The third bypasses explicit rewards entirely, instead learning directly from preference data, as in Diffusion-DPO~\citep{wallace2024diffusion} and Diffusion-KTO~\citep{yang2024using}. More recently, GRPO has been successfully adapted to diffusion models~\citep{flowgrpo,dancegrpo,dgpo}, showing strong scalability and notable performance gains. Nevertheless, all these approaches target standard diffusion models. Our work tackles a distinct problem: reinforcing the few-step diffusion models.

\spara{Few-Step Text-to-Image Diffusion RL} As for few-step generative models, recent work~\citep{luo2024diff,luo2024diffstar,li2024reward, ren2024hyper, luo2025reward} has introduced methods for training preference-aligned few-step text-to-image models, achieving promising results. Nevertheless, these existing approaches for few-step models are restricted to differentiable reward functions, with narrow applications in cases of non-differentiable rewards. Some others theoretically support non-differentiable rewards~\cite{pso,dmdr} by combining standard RL methods, but they have not validated their effectiveness at scales. Furthermore, applying standard RL methods of diffusion models to few-step models would lead to blurry generation and suboptimal performance, as illustrated in \cref{fig:compare_baseline,fig:compare_niave}. To our best knowledge, our TDM-R1 is the first work tackling RL with non-differentiable rewards for few-step text-to-image generative models at scales.

\section{Conclusion}
In this paper, we introduced TDM-R1, a novel reinforcement learning paradigm that enables few-step diffusion models to effectively leverage non-differentiable reward feedback. By building upon Trajectory Distribution Matching and decoupling the learning process into surrogate reward learning and generator optimization, TDM-R1 addresses a fundamental limitation of existing approaches that rely exclusively on differentiable reward signals. Our extensive experiments demonstrate that TDM-R1 consistently achieves state-of-the-art reinforcement learning performance for few-step text-to-image models. Most notably, TDM-R1 enables 4-step models to surpass their expensive 80-NFE base counterparts, with improvements on GenEval from 61\% to 92\%—exceeding both the 80-NFE base model at 63\%. These results validate that few-step diffusion models can successfully incorporate large-scale online non-differentiable reward signals without requiring additional ground-truth image data.

\bibliography{refs}
\appendix
\setcounter{tocdepth}{2}
\allowdisplaybreaks
\newpage
\newpage
\section{Derivation}
\label{app:derivation}

\subsection{Derivation of \cref{eq:kl_dgpo}}
For simplicity and without loss of generality, we omit the conditioning on $\rvc$ throughout this derivation. By substituting the surrogate reward parameterization from \cref{eq:noisy_reward} into the Bradley-Terry objective in \cref{eq:logp_bt}, we obtain the initial training objective:
\begin{equation}
\label{eq:initial_objective}
\begin{aligned}
L(\phi) &= -\E_k \E_{(\mathcal{G}_k^+, \mathcal{G}_k^-) \sim \mathcal{D}} \log\sigma\Bigg( \sum_{\rvx_{t_k} \in \mathcal{G}_k^+} \E_{q(\rvx_{t_k+1:T}|\rvx_{t_k})} \beta w(\rvx_{t_k}) \left[ \log \frac{p_\phi(\rvx_{t_k:T})}{p_{\text{ref}}(\rvx_{t_k:T})} + \log Z \right] \\
&\quad - \sum_{\rvx_{t_k} \in \mathcal{G}_k^-} \E_{q(\rvx_{t_k+1:T}|\rvx_{t_k})} \beta w(\rvx_{t_k}) \left[ \log \frac{p_\phi(\rvx_{t_k:T})}{p_{\text{ref}}(\rvx_{t_k:T})} + \log Z \right] \Bigg).
\end{aligned}
\end{equation}

The partition function terms cancel because the weights sum to zero over the full group. Specifically, since the weights are defined as absolute normalized advantages and the partition into positive and negative groups is based on the sign of the advantage, we have $\sum_{\rvx_{t_k} \in \mathcal{G}_k^+} w(\rvx_{t_k}) - \sum_{\rvx_{t_k} \in \mathcal{G}_k^-} w(\rvx_{t_k}) = \sum_{\rvx_{t_k} \in \mathcal{G}_k} A(\rvx_{t_k})=  \sum_{\rvx_{t_k} \in \mathcal{G}_k} \frac{r(\rvx_k) - r_{mean}}{r_{std}} = 0$, where $r_{mean} = \frac{\sum_{\rvx_{t_k} \in \mathcal{G}_k} r(\rvx_k)}{|G|}$. This simplifies the objective to:
\begin{equation}
\label{eq:simplified_objective}
\begin{aligned}
L(\phi) &= -\E_{(\mathcal{G}^+, \mathcal{G}^-) \sim \mathcal{D}} \log\sigma\Bigg( \beta \bigg[ \sum_{\rvx_{t_k} \in \mathcal{G}^+} w(\rvx_{t_k}) \E_{q(\rvx_{t_k+1:T}|\rvx_{t_k})} \log \frac{p_\phi(\rvx_{t_k:T})}{p_{\text{ref}}(\rvx_{t_k:T})} \\
&\quad - \sum_{\rvx_{t_k} \in \mathcal{G}^-} w(\rvx_{t_k}) \E_{q(\rvx_{t_k+1:T}|\rvx_{t_k})} \log \frac{p_\phi(\rvx_{t_k:T})}{p_{\text{ref}}(\rvx_{t_k:T})} \bigg] \Bigg).
\end{aligned}
\end{equation}

Next, we decompose the trajectory-level log-ratio using the Markov property of the diffusion process. For any trajectory $\rvx_{t_k:T}$, the joint distribution factorizes as a product of transition probabilities:
\begin{equation}
\label{eq:markov_factorization}
\log \frac{p_\phi(\rvx_{t_k:T})}{p_{\text{ref}}(\rvx_{t_k:T})} = \sum_{t=t_k+1}^{T} \log \frac{p_\phi(\rvx_{t-1}|\rvx_t)}{p_{\text{ref}}(\rvx_{t-1}|\rvx_t)}.
\end{equation}

Substituting this factorization and rewriting the sum over timesteps as an expectation, we obtain:
\begin{equation}
\label{eq:timestep_expectation}
\begin{aligned}
L(\phi) &= -\E_{(\mathcal{G}^+, \mathcal{G}^-) \sim \mathcal{D}} \log\sigma\Bigg( \beta(T-t_k) \E_{t \sim \mathcal{U}[t_k+1, T]} \E_{q(\rvx_t|\rvx_{t_k}), q(\rvx_{t-1}|\rvx_t, \rvx_{t_k})} \\
&\quad \times \bigg[ \sum_{\rvx_{t_k} \in \mathcal{G}^+} w(\rvx_{t_k}) \log \frac{p_\phi(\rvx_{t-1}|\rvx_t)}{p_{\text{ref}}(\rvx_{t-1}|\rvx_t)} - \sum_{\rvx_{t_k} \in \mathcal{G}^-} w(\rvx_{t_k}) \log \frac{p_\phi(\rvx_{t-1}|\rvx_t)}{p_{\text{ref}}(\rvx_{t-1}|\rvx_t)} \bigg] \Bigg).
\end{aligned}
\end{equation}

To obtain a tractable upper bound, we apply Jensen's inequality. Since $-\log\sigma(\cdot)$ is a convex function, moving the expectation over $t$ and $q(\rvx_t|\rvx_{t_k})$ inside the sigmoid yields:
\begin{equation}
\label{eq:jensen_bound}
\begin{aligned}
L(\phi) &\leq \E_{(\mathcal{G}^+, \mathcal{G}^-) \sim \mathcal{D}} \E_{t, q(\rvx_t|\rvx_{t_k})} \Bigg[ -\log\sigma\bigg( \beta(T-t_k) \E_{q(\rvx_{t-1}|\rvx_t, \rvx_{t_k})} \\
&\quad \times \Big[ \sum_{\rvx_{t_k} \in \mathcal{G}^+} w(\rvx_{t_k}) \log \frac{p_\phi(\rvx_{t-1}|\rvx_t)}{p_{\text{ref}}(\rvx_{t-1}|\rvx_t)} - \sum_{\rvx_{t_k} \in \mathcal{G}^-} w(\rvx_{t_k}) \log \frac{p_\phi(\rvx_{t-1}|\rvx_t)}{p_{\text{ref}}(\rvx_{t-1}|\rvx_t)} \Big] \bigg) \Bigg].
\end{aligned}
\end{equation}

Finally, we rewrite the log-ratios in terms of KL divergences. For any distribution $q$ and models $p_\phi$, $p_{\text{ref}}$, we have the following identity:
\begin{equation}
\label{eq:kl_identity}
\E_{q(\rvx_{t-1}|\rvx_t, \rvx_{t_k})} \log \frac{p_\phi(\rvx_{t-1}|\rvx_t)}{p_{\text{ref}}(\rvx_{t-1}|\rvx_t)} = \kl\big(q(\rvx_{t-1}|\rvx_t, \rvx_{t_k}) \| p_{\text{ref}}(\rvx_{t-1}|\rvx_t)\big) - \kl\big(q(\rvx_{t-1}|\rvx_t, \rvx_{t_k}) \| p_\phi(\rvx_{t-1}|\rvx_t)\big).
\end{equation}

Substituting this identity and rearranging terms, we arrive at the final upper bound:
\begin{equation}
\label{eq:final_bound}
\begin{aligned}
L(\phi) &\leq \E_{(\mathcal{G}_k^+, \mathcal{G}_k^-) \sim \mathcal{D}_k, k} \E_{t, q(\rvx_t|\rvx_{t_k})} \log\sigma\Bigg( -\beta(T-t_k) \bigg\{ \\
&\quad \sum_{\rvx_{t_k} \in \mathcal{G}_k^+} w(\rvx_{t_k}) \Big[ \kl\big(q(\rvx_{t-1}|\rvx_t, \rvx_{t_k}) \| p_\phi(\rvx_{t-1}|\rvx_t)\big) - \kl\big(q(\rvx_{t-1}|\rvx_t, \rvx_{t_k}) \| p_{\text{ref}}(\rvx_{t-1}|\rvx_t)\big) \Big] \\
&\quad - \sum_{\rvx_{t_k} \in \mathcal{G}_k^-} w(\rvx_{t_k}) \Big[ \kl\big(q(\rvx_{t-1}|\rvx_t, \rvx_{t_k}) \| p_\phi(\rvx_{t-1}|\rvx_t)\big) - \kl\big(q(\rvx_{t-1}|\rvx_t, \rvx_{t_k}) \| p_{\text{ref}}(\rvx_{t-1}|\rvx_t)\big) \Big] \bigg\} \Bigg).
\end{aligned}
\end{equation}

This completes the derivation of \cref{eq:kl_dgpo}.

\subsection{Derivation of \cref{eq:tdm_r1_grad}}
We compute the gradient of the objective in \cref{eq:k_rlhf} by separately deriving the gradients of the reward maximization and KL regularization terms.

\textbf{Step 1: Reformulating the Surrogate Reward.}
We first re-write the surrogate reward defined in \cref{eq:noisy_reward} as follows:
\begin{equation}
\label{eq:noisy_reward_v2}
\begin{aligned}
    \Tilde{r}_\phi(\rvx_{t_k}, \rvc) 
    & = \beta \E_{q(\rvx_{t_{k+1}:T}|\rvx_{t_k})} \log \frac{p_\phi(\rvx_{t_k:T}|\rvc)}{p_{\text{ref}}(\rvx_{t_k:T}|\rvc)} + \beta\log Z \\
    & = \beta \E_{q(\rvx_{t_{k+1}:T}|\rvx_{t_k})} \sum_{t=t_k + 1}^{T} \log \frac{p_\phi(\rvx_{t-1}|\rvx_{t})}{p_{\text{ref}}(\rvx_{t-1}|\rvx_{t})} + \beta \log Z.
\end{aligned}
\end{equation}
By converting the summation over timesteps into an expectation with respect to a uniform distribution over $t \geq t_k$, we obtain:
\begin{equation}
\begin{aligned}
    \Tilde{r}_\phi(\rvx_{t_k}, \rvc) 
    & = \beta (T - t_k) \E_{t \geq t_k} \E_{q(\rvx_{t_{k+1}:T}|\rvx_{t_k})} \log \frac{p_\phi(\rvx_{t-1}|\rvx_{t})}{p_{\text{ref}}(\rvx_{t-1}|\rvx_{t})} + \beta \log Z \\
    & = \beta (T - t_k) \E_{t \geq t_k} \E_{q(\rvx_{t}|\rvx_{t_k})} \E_{q(\rvx_{t-1}|\rvx_{t}, \rvx_{t_k})} \log \frac{p_\phi(\rvx_{t-1}|\rvx_{t})}{p_{\text{ref}}(\rvx_{t-1}|\rvx_{t})} + \beta \log Z,
\end{aligned}
\end{equation}
where the last equality follows from marginalizing the joint distribution $q(\rvx_{t_{k+1}:T}|\rvx_{t_k})$ to the relevant variables $\rvx_t$ and $\rvx_{t-1}$.

\textbf{Step 2: Gradient of the Reward Term.}
Applying the chain rule through the reparameterized sampling process, the gradient of the reward term with respect to $\theta$ is given by:
\begin{equation}
\label{eq:r_grad}
\begin{aligned}
    \nabla_\theta \E_{k, p_{\theta}(\rvx_{t_k})} \Tilde{r}_\phi(\rvx_{t_k}, \rvc) 
    & = \beta (T - t_k) \E_{k, t \geq t_k} \E_{p_{\theta}(\rvx_{t_k})} \E_{q(\rvx_t, \rvx_{t-1}|\rvx_{t_k})} \left[ \nabla_{\rvx_t} \log \frac{p_\phi(\rvx_{t-1}|\rvx_{t})}{p_{\text{ref}}(\rvx_{t-1}|\rvx_{t})} \right] \frac{\partial \rvx_t}{\partial \rvx_{t_k}} \frac{\partial \rvx_{t_k}}{\partial \theta}.
\end{aligned}
\end{equation}
Since the forward diffusion process $q(\rvx_t | \rvx_{t_k})$ is Gaussian with $\rvx_t = \alpha_{t|t_k} \rvx_{t_k} + \sigma_{t|t_k} \boldsymbol{\epsilon}$, where $\boldsymbol{\epsilon} \sim \mathcal{N}(\mathbf{0}, \mathbf{I})$, we have $\frac{\partial \rvx_t}{\partial \rvx_{t_k}} = \alpha_{t|t_k}$. Substituting this yields:
\begin{equation}
\begin{aligned}
    \nabla_\theta \E_{k, p_{\theta}(\rvx_{t_k})} \Tilde{r}_\phi(\rvx_{t_k}, \rvc) 
    & = \beta (T - t_k) \E_{k, t \geq t_k} \E_{p_{\theta}(\rvx_{t_k})} \E_{q(\rvx_t, \rvx_{t-1}|\rvx_{t_k})} \left[ \alpha_{t|t_k} \nabla_{\rvx_t} \log \frac{p_\phi(\rvx_{t-1}|\rvx_{t})}{p_{\text{ref}}(\rvx_{t-1}|\rvx_{t})} \right] \frac{\partial \rvx_{t_k}}{\partial \theta}.
\end{aligned}
\end{equation}
Note that the gradient with respect to $\log Z$ vanishes since the partition function is independent of $\rvx_t$.

\textbf{Step 3: Gradient of the KL Regularization Term.}
The gradient of the marginal-level KL divergence term follows from \cref{eq:tdm_grad}.

\textbf{Step 4: Combining the Gradients.}
Combining the results from Steps 2 and 3, and incorporating the negative sign from the reward maximization term in \cref{eq:k_rlhf}, we arrive at the final gradient expression:
\begin{equation}
\begin{aligned}
    \nabla_\theta L(\theta) & = -\E_{k, t \geq t_k} \E_{p_{\theta}(\rvx_{t_k})} \E_{q(\rvx_t, \rvx_{t-1}|\rvx_{t_k})} \Bigg[ \\ 
    & \quad \beta \alpha_{t|t_k} (T - t_k) \nabla_{\rvx_{t}} \log \frac{p_\phi(\rvx_{t-1}|\rvx_{t})}{p_{\text{ref}}(\rvx_{t-1}|\rvx_{t})} \\ 
    & \quad + \beta_g \lambda_t \left( s_{\text{fake}}(\rvx_t) - s_\psi(\rvx_t) \right) \Bigg] \frac{\partial \rvx_{t_{k}}}{\partial \theta},
\end{aligned}
\end{equation}
which completes the derivation of \cref{eq:tdm_r1_grad}.

\section{Experiment details} 
\label{app:exp_detail}

\paragraph{Visual Text Rendering}
Adopting the evaluation protocol from TextDiffuser~\citep{chen2023textdiffuser} and the experimental framework of Flow-GRPO, we assess how accurately models can render textual content within generated images. Prompts follow a standardized template: ``A sign that says `text''', where `text' denotes the target string to be visually rendered. Text fidelity~\citep{gong2025seedream} is computed as:
\[
r = \max(1 - N_e / N_{\text{ref}}, 0)
\]
Here, $N_e$ represents the minimum edit distance between the rendered output and the target text, while $N_{\text{ref}}$ indicates the character length of the quoted string in the prompt.

\paragraph{Setup Details} 
During training, we generate 24 samples per group. We employ Flow-DPM-Solver~\citep{xie2025sana} with 4 sampling steps for both training rollouts and inference. We apply LoRA fine-tuning with rank 32, and set $\beta = 100$ as the default. Our default choice of $\beta_g$ is set such that the ratio of the gradient of the reward term to the gradient of the KL regularization term is $2:1$. All experiments are conducted at 512 resolution, with GPU hours reported in A100 equivalents.

\paragraph{Details of the out-of-domain evaluation metrics}
We describe the specific out-of-domain metrics employed for quality assessment. The \texttt{aesthetic score}~\citep{schuhmann2022laion} uses a CLIP-based linear regression model to quantify the visual appeal of generated images. To evaluate image quality degradation, we employ the \texttt{DeQA score}~\citep{deqa}, which utilizes a multimodal large language model architecture to measure how various imperfections—such as distortions, textural degradation, and low-level visual artifacts—affect overall perceived image quality.
\texttt{ImageReward}~\citep{kirstain2023pick} assesses both visual quality and text-image correspondence, offering a holistic evaluation of generation quality from a human-centered standpoint.
\texttt{ImageReward}~\citep{xu2023imagereward} functions as a comprehensive human preference model for text-to-image generation, evaluating multiple aspects including text-visual coherence and generation fidelity. Lastly, \texttt{UnifiedReward}~\citep{wang2025unified} represents a state-of-the-art advancement in this domain. This unified reward framework is capable of evaluating both multimodal understanding and generation tasks, and has achieved superior performance on human preference assessment benchmarks compared to prior methods.

\section{Additional Qualitative Comparison}
\label{app:add_samples}

\begin{figure}[t]
    \centering
    \includegraphics[width=1\linewidth]{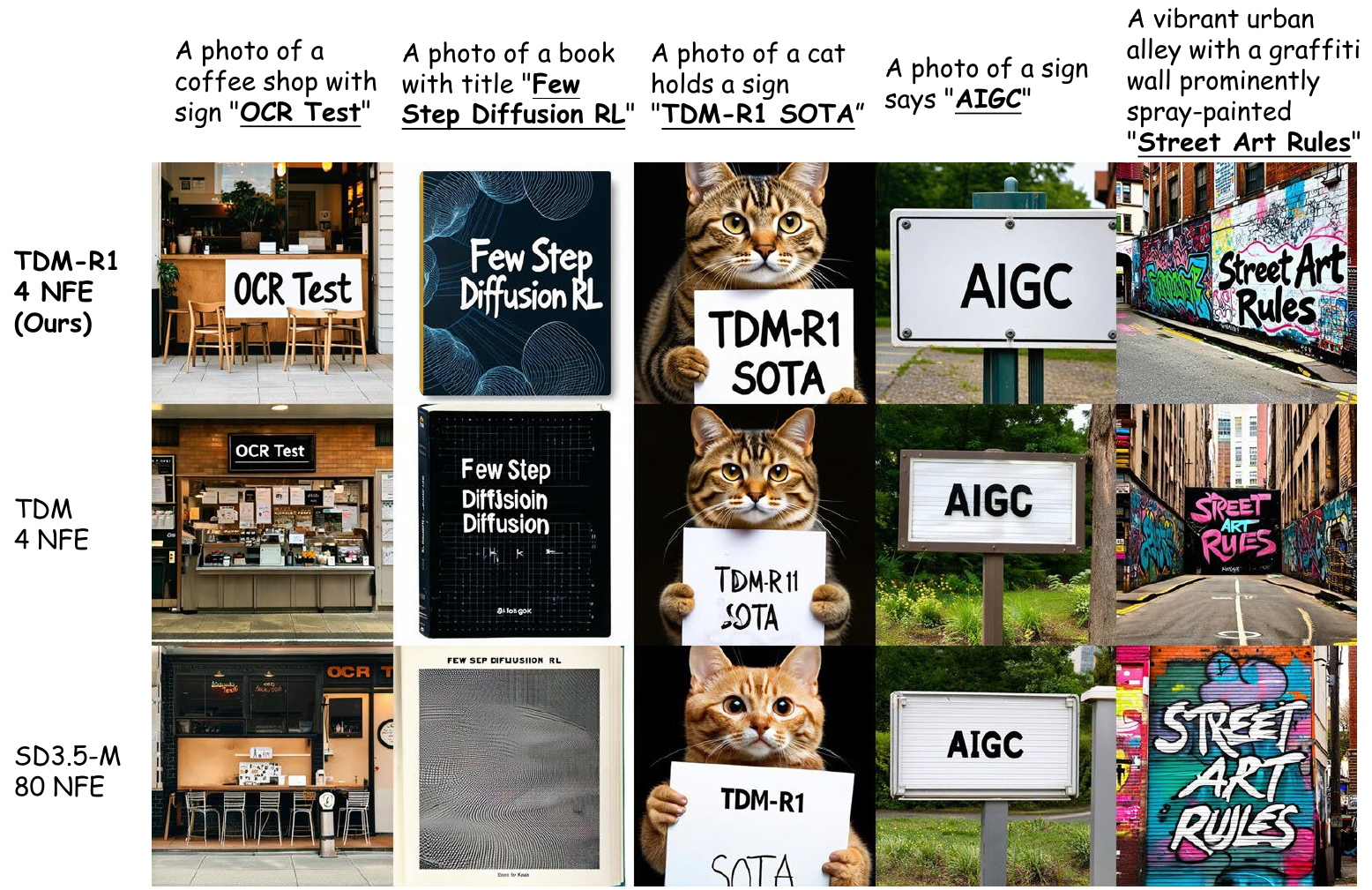}
    \vspace*{-4mm}
    \caption{Qualitative comparisons of TDM-R1 against competing baselines. The same initial noise is used to generate all images.}
    \label{fig:visual_compare_ocr}
\end{figure}

We present additional visual comparison in \cref{fig:visual_compare_ocr}.

\end{document}